\documentclass{article}
\usepackage[utf8]{inputenc}

\title{Energy Efficient Hadamard Neural Networks}
\author{T. Ceren Deveci, Serdar Cakir and A. Enis Cetin}
\date{April 2018}

\usepackage{cite}
\usepackage{graphicx}
\usepackage[usenames]{color} 
\usepackage{amssymb} 
\usepackage{graphicx}
\usepackage{amsmath,amsfonts,amsthm,bm} 
\usepackage{url}
\usepackage{array} 
\usepackage{multirow}
\usepackage{tabularx}
\usepackage{algorithm} 
\usepackage{algcompatible} 
\usepackage{algpseudocode}
\usepackage{cite} 
\usepackage{hyperref} 
\usepackage{subfigure}

\begin{document}

\maketitle

\begin{abstract}
Deep learning has made significant improvements at many image processing tasks in recent years, such as image classification, object recognition and object detection. Convolutional neural networks (CNN), which is a popular deep learning architecture designed to process data in multiple array form, show great success to almost all detection \& recognition problems and computer vision tasks. However, the number of parameters in a CNN is too high such that the computers require more energy and larger memory size. In order to solve this problem, we propose a novel energy efficient model Binary Weight and Hadamard-transformed Image Network (BWHIN), which is a combination of Binary Weight Network (BWN) and Hadamard-transformed Image Network (HIN). It is observed that energy efficiency is achieved with a slight sacrifice at classification accuracy. Among all energy efficient networks, our novel ensemble model outperforms other energy efficient models.
\end{abstract}

\section{Introduction}

In the recent years, artificial neural networks (ANN) have become very popular amongst researchers due to their great success on image classification, feature extraction, segmentation, object recognition and detection~\cite{egmont2002image}. Deep learning is a more sophisticated and particular form of machine learning, which enables the user to form complex models which are composed of multiple hidden layers. Deep learning methods have enhanced the state-of-the-art performance in object recognition \& detection and computer vision tasks. Deep learning is also advantageous for processing raw data such that it can automatically find a suitable representation for detection or classification~\cite{lecun2015deeplearning-nature}.

Convolutional neural network (CNN) is a specific deep learning architecture for processing data which is composed of multiple arrays. Images can be a good example of input to CNN with its 2D grid of pixels. Convolutional Neural Networks have become popular with the introduction of its modern version \textit{LeNet-5} for the recognition of handwritten numbers~\cite{lecun1998gradient-lenet5}. Besides, \textit{AlexNet}, the winner of ILSVRC object recognition challenge in 2012, aroused both commercial and scientific interest in CNN and it is the main reason of the intense popularity of CNN architectures for deep learning applications~\cite{krizhevsky2012imagenet-alexnet}. The usage of CNN in AlexNet obtained remarkable results such that the network halved the error rate of its previous competitors. Thanks to this great achievement, CNN is the most preferred approach for most detection and recognition problems and computer vision tasks. 

Although CNNs are suitable for efficient hardware implementations such as in GPUs or FPGAs, the training is computationally expensive due to the high number of parameters. As a result, excessive amount of energy consumption and memory usage make the implementation of neural networks ineffective. According to ~\cite{han2015learning}, especially matrix multiplications at the layers of a neural network consume too much energy compared to addition or activation function and becomes a major problem for mobile devices with limited batteries. As a result, replacing the multiplication operation becomes the main concern in order to achieve energy efficiency. 

Many solutions are proposed in order to handle the energy efficiency problem. An energy efficient $\ell_{1}$-norm based operator is introduced in~\cite{tuna2009image-multiplierless}. This multiplier-less operator is first used in image processing tasks such as cancer cell detection and licence plate recognition in~\cite{akbas2014l1-multiplierfree,suhre2013multiplication}. Multiplication-free neural networks (MFNN) based on this operator are studied in~\cite{akbacs2015multiplication,afrasiyabi2017energy,badawi2017multiplication}. This operator achieved promising performance especially at image classification on MNIST dataset with multi-layer perceptron (MLP) models~\cite{afrasiyabi2017energy}. Han et al. reduces both the computation and storage in three steps: First, the network is trained to learn the important connections. Then, the redundant connections are discarded for a sparser network. Finally, the remaining network is retrained~\cite{han2015learning}. Using neuromorphic processors with its special chip architecture is another solution for energy efficiency~\cite{esser2016convolutional}. In order to improve energy consumption, Sarwar et al. exploits the error resiliency of artificial neurons and approximates the multiplication operation and defines a Multiplier-less Artificial Neuron (MAN) by using Alphabet Set Multiplier (ASM). In ASM, the multiplication is approximated as shifting and adding in bitwise manner with some previously defined alphabets~\cite{sarwar2016multiplier}. Binary Weight Networks are energy efficient neural networks whose filters at the convolutional layers are approximated as binary weights. With these binary weights, convolution operation can be computed only with addition and subtraction~\cite{rastegari2016xnor-binaryweightnetwork}. There is also a computationally inexpensive method called distillation~\cite{hinton2015distilling}. A very large network or an emsemble model is first trained and transfers its knowledge to a much smaller, \textit{distilled} network. Using this small and compact model is much more advantageous in mobile devices in terms of speed and memory size. This method shows promising results at image processing tasks such as facial expression recognition~\cite{ccuugu2017microexpnet}.

In this study, an energy-efficient neural network framework based on Binary Weight Network (BWN)~\cite{rastegari2016xnor-binaryweightnetwork} and Hadamard Transform is developed. The weights at the convolutional layers of the BWN are approximated to binary values, $+1$ or $-1$~\cite{rastegari2016xnor-binaryweightnetwork}. Instead of utilizing the original images as network inputs, the network is modified to use compressed images. This network is called Hadamard-transformed Image Network (HIN). Since Hadamard transform is implemented by Fast Walsh-Hadamard Transform algorithm which requires only addition or subtraction~\cite{fino1976unified}, the HIN network is energy efficient. Our main contribution is the combination of BWN and HIN models: Binary Weight and Hadamard-transformed Image Network (BWHIN). The combination is carried out after the energy efficient layers, i.e. convolutional layers with two different averaging techniques. All of the energy efficient models are also examined with different CNN architectures. One of them (ConvoPool-CNN) contains pooling layers along with convolutional layers, while the other (All-CNN~\cite{springenberg2014striving-allconvnet}) uses strided convolution instead of pooling layer~\cite{springenberg2014striving-allconvnet}. We analyze the performance of the models on two famous image datasets MNIST and CIFAR-10. While working on MNIST, we also study the effects of certain hyperparameters on the classification accuracy of energy efficient neural networks.

\section{Methodology}

Firstly, we investigate the efficiency BWN proposed in~\cite{rastegari2016xnor-binaryweightnetwork} which approximates the weights to binary values. Similar to BWN, we propose a Hadamard-transformed Image Network (HIN) which uses the Hadamard-transformed images with binarized weights. Lastly, a combined network is introduced and its performance is compared with BWN and HIN frameworks.

\subsection{Binary Weight Networks (BWN)} \label{ch_bwn}

Binary-Weight Network (BWN) is proposed in~\cite{rastegari2016xnor-binaryweightnetwork} as an efficient approximation to standard convolutional neural networks. In BWNs, the filters, i.e. weights of the CNN are approximated to binary values $+1$ and $-1$. While a conventional convolutional neural network needs multiplication, addition and subtraction for convolution operation, convolution with binary weights can be estimated by only addition and subtraction. \par

Convolution operation can be approximated as $
\bm{I} \ast \bm{W} 	\approx (\bm{I} \oplus \bm{B}) \alpha $
where $\bm{I}$ is the input tensor, $\bm{W}$ is the weight (filter), $\bm{B}$ is the binary weight tensor which has the same size with $\bm{W}$ and $\alpha \in \mathbb{R}^{+}$ is the scaling factor such that $\bm{W} \approx \alpha \bm{B}$. $\oplus$ operation indicates convolution only with addition and subtraction. Since the weight values are only $+1$ and $-1$, convolution operation can be implemented in a multiplier-less manner. After solving an optimization problem to estimate $\bm{W}$, $\bm{B}$ and $\alpha$ is found as: \par

\begin{equation} \label{b-binary weight}
\bm{B}=sign(\bm{W})
\end{equation}

\begin{equation} \label{alpha-binary weight}
\alpha=\dfrac{\sum_{}^{} |\bm{W}_{i}|}{n}=\dfrac{1}{n} ||\bm{W}||_{\ell 1}
\end{equation}

In Equation \ref{alpha-binary weight}, $n=c \times w \times h$ where $c$ is the channel, $h$ is the height and $w$ is the width of weight tensor $\bm{W}$, and of $\bm{B}$ as well. Equations \ref{b-binary weight} and \ref{alpha-binary weight} show that binary weight filter is simply the sign of weight values and scaling factor is the average of absolute weight values. While training a CNN with binary weights, the weights are only binarized in forward pass and back propagation steps of the training. At the parameter-update stage, the real-valued weights (not binarized) are used. Another significant point about this network is that convolutional filters here don't have bias terms, and this convolution approximation is only held in convolutional layers. Fully connected layers still do have bias terms and standard multiplication.

\subsection{Hadamard-transformed Image Networks (HIN)}

In the literature, transform domain features are also used as input data in deep learning structures. Although feature extraction requires extra computation time and energy, transform domain features as input to the convolutional neural networks can be preferred due to simpler implementation, effective computation and improved model accuracy. Discrete Cosine Transform (DCT) domain data as the input data can outperform the state-of-the-art results as shown in ~\cite{wu2017compressed}. Wu et al. uses DCT because JPEG, MPEG video coding standards are based on DCT. In this study, we also use transform domain features to feed them into the our CNN model. This network model is called Hadamard-transformed Image Networks (HIN). \par

Compressed domain data is also used as input in deep learning structures. Discrete Cosine Transform (DCT) domain data as the input data can outperform the state-of-the-art results as shown in ~\cite{wu2017compressed}. Compressed domain video frames as input to the convolutional neural networks are preferred rather than RGB frames, since data decompression requires extra computation time and energy. As a result; simpler implementation, effective computation and improved model accuracy are achieved. Wu et al. uses DCT because JPEG, MPEG video coding standards are based on DCT. In this study, we also use transform domain features to feed them into the our CNN model. This network model is called Hadamard-transformed Image Networks (HIN). \par

Hadamard Transform, also called as Hadamard-ordered Walsh-Hadamard Transform, is an image transformation technique which is also used to compress images~\cite{petrou2010image}. Hadamard Transform coefficients consists of binary values +1 and -1. Thus, Hadamard Transform can be considered as an efficient alternative to the other image transforms such that it can be implemented without any multiplication and division~\cite{rafael2007digital}.

1-D Hadamard Transform is expressed by Equation~\ref{1d-hadamard-formula}. In this formula, $g(x)$ is the elements of 1-D array $\underline{g}$ and $b_{i}(x)$ is the $i^{th}$ bit (from right to left) in the binary representation of $x$. The scaling factor $(\dfrac{1}{\sqrt{2}})^{m}$ is used to make the Hadamard matrix orthonormal, hence it is mostly kept in the calculations. \par

\begin{equation} \label{1d-hadamard-formula}
T(u)=(\dfrac{1}{\sqrt{2}})^{m} \sum_{x=0}^{2^{m}-1} g(x) (-1)^{\sum\limits_{i=0}^{m-1} b_{i}(x)b_{m-1-i}(u)}
\end{equation}

2-D Hadamard Transform is a straightforward extension of 1-D Hadamard Transform~\cite{pratt1969hadamard}: \par

\begin{equation} \label{2d-hadamard-formula}
T(u,v)=(\dfrac{1}{2})^{m} \sum_{x=0}^{2^{m}-1} \sum_{y=0}^{2^{m}-1} g(x,y) (-1)^{\sum\limits_{i=0}^{m-1} (b_{i}(x)p_{i}(u)+b_{i}(y)p_{i}(v))}
\end{equation}

In Equation~\ref{2d-hadamard-formula}, $p_{i}(u)$ is computed using: 
\small
\begin{equation}
\begin{split}
p_{0}(u) & = b_{m-1}(u) \\
p_{1}(u) & = b_{m-1}(u) + b_{m-2}(u) \\
\vdots \\
p_{m-1}(u) & = b_{1}(u) + b_{0}(u) \\
\end{split}
\end{equation}
\normalsize

2-D Hadamard Transform is separable and symmetric, hence it can be implemented by using row-column or column-row passes of the corresponding 1-D transform. \par

There is an algorithm called Fast Walsh-Hadamard Transform ($FWTH_{h}$) which requires less storage and is fast and efficient to compute Hadamard Transform~\cite{fino1976unified}. The implementation of this algorithm can be realized by only addition and subtraction operations which can be summarized in a butterfly structure. While the complexity of Hadamard Transform is $O(N^{2})$, complexity of fast algorithm is $O(Nlog_{2}N)$ where $N=2^{m}$.If the length of the input 1-D array is less than a power of 2, the array is padded with zeros up to the next greater power of two. Since 2-D Hadamard Transform is separable, we can treat columns and rows of the 2-D array as separate 1-D arrays.

Training of HIN is similar to the training of BWN; the only difference is that the input images are Hadamard-transformed as explained above. Training proceeds as explained in Section \ref{ch_bwn}, but at the beginning Hadamard-transformed input data is fed in to the network instead. As in BWN, binarized weights are used and no bias terms are defined.

\subsection{Combination of Models: Binary Weight \& Hadamard Transformed Image Network (BWHIN)} \label{combin-ensemble}

Combination of the neural networks can improve the performance of the neural networks by a few percent. Since combining the neural networks reduces the test error and tends to keep the training error the same, it can be viewed as a regularization technique. One of the popular techniques of the combination is called ``model ensembles" which combines the multiple hypotheses that explain the same training data~\cite{goodfellow2016deep,cs231n-learningrate}. In model ensembles, the error made by averaging prediction of all models in the ensemble decreases linearly with the ensemble size, i.e. the number of models in the ensemble. However, since they need longer time and higher amount of memory to evaluate on test example, we try to avoid increasing the ensemble size for energy efficiency. In multi-network frameworks, different networks are trained independently and separately before performing a combination layer. Bilinear CNNs~\cite{lin2015bilinearcnn} is a good example for such combination models. In bilinear CNN, there are two sub-networks which are standard CNN units. After these CNN units, the image regions which extract features are combined with a matrix dot product and then average pooled to obtain the bilinear vector. In order to perform these operations properly, those image regions have to be of the same size. This vector is passed through a fully-connected and softmax layer to obtain class predictions. \par

\begin{figure}
	\centering
	\includegraphics[width=10.5cm]{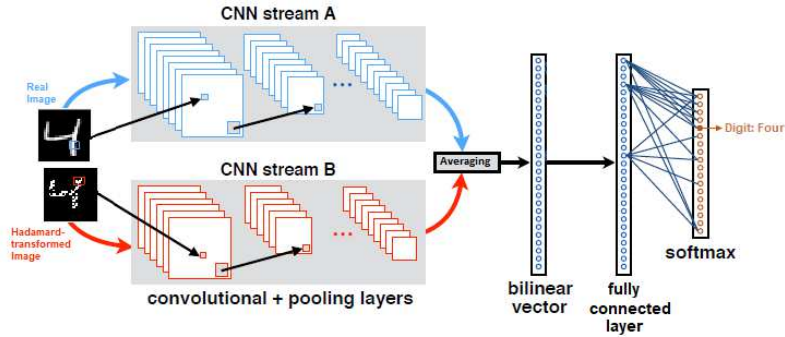}
	\caption{Our approach to combine BWN and HIN: The architecture of BWHIN~\cite{lin2015bilinearcnn}.}
	\label{fig:combinationcnn}
\end{figure}

Our approach to combine BWN and HIN is quite similar to Bilinear-CNN, but simpler. After convolutional, ReLU and pooling layers, the output tensor is reshaped for fully connected layer as a 1-D tensor. Afterwards, these same sized 1-D tensors of each sub-network will be averaged instead of dot product. Since multiplication consumes power, dot product is avoided and averaging is preferred. Two averaging methods are used: Simple averaging and weighted averaging.~\cite{yang2004neural}. Simple averaging is the conventional averaging technique which calculates the output by averaging the sum of outputs from each ensemble member. Weighted averaging technique assigns a weight to each ensemble member and calculated the output by taking these weights into account. The total weight of each ensemble is 1. In order to implement this technique, we define a random number which behaves like a weight. If $YY_{binary}$ is defined as the 1-D tensor of BWN and $YY_{hadamard}$ is the 1-D tensor of HIN, the weighted averaging is defines as follows:

\begin{equation} \label{randomaverage-eqn}
YY_{combined}=(W_{combined} \times YY_{binary})  + ((1-W_{combined})) \times YY_{hadamard}
\end{equation}
where $W_{combined}$ is initialized according to truncated normal distribution which can only take values in $\left[0, 1\right]$. After the averaging operation, the resultant 1-D tensor is processed through a fully connected and softmax layers. The architecture of our combined network Binary Weight \& Hadamard-Transformed Image Network (BWHIN) is summarized in Figure~\ref{fig:combinationcnn}. 

By looking at Figure~\ref{fig:combinationcnn}, one can observe that the combination is applied after the convolutional layers of each network, which are energy efficient layers. With this combination model, we still want to maintain the energy efficiency of the entire network.

\section{Experimental Studies}

\subsection{Image Datasets and Implementation}

We analyze our proposed algorithm on two well-known datasets: MNIST and CIFAR-10.
The MNIST (Modified National Institute of Standards and Technology) database of handwritten digit images is a very popular digit database for implementing learning techniques and pattern recognition methods~\cite{lecun1998mnist}. It contains 60,000 training images and 10,000 test images in ten classes. These black and white images are of size $28\times28$ pixels.

CIFAR-10 (Canadian Institute for Advanced Research-10) is also a popular dataset used for image classification tasks~\cite{krizhevsky2009learning-cifar10}. It consists of 50,000 training and 10,000 test images. These images, which are of size $32\times32$, are collected in ten different classes of objects (airplane, automobile, bird, cat, deer, dog, frog, horse, ship, truck).

Tensorflow is chosen to implement all of the deep neural networks for this work. Tensorflow is an open-source software library for machine intelligence, which is developed by Google Brain Team in Python language~\cite{abadi2016tensorflow}. Since our main purpose is to investigate the energy efficieny of the proposed neural network models, simple architectures are chosen and no pretraining such as feature extraction or unsupervised learning techniques is performed. Images in both datasets are normalized such that the pixel values are in $\left[0, 1\right]$. All experiments are carried out on a single GPU, which is NVIDIA GeForce 940M. Thanks to our CUDA-enabled GPU, we are able to run Tensorflow with GPU support and we achieve faster computation. \par

\subsection{Neural Network Architecture and Hyperparamaters} \label{hyperselect}

\subsubsection{CNN Architectures}

In order to implement the proposed networks and analyze their performances, two different CNN architectures are utilized. First type of architecture for MNIST database is very similar to LeNet-5 in~\cite{lecun1998gradient-lenet5} with convolutional and pooling layers. Second architecture is built according to All-Convolutional-Neural-Network~\cite{springenberg2014striving-allconvnet} with strided convolution. Strided convolution is that some positions of the kernel are skipped over in order to reduce the computational burden while implementing the convolution operation. Strided convolution is equivalent to downsampling the output of the full convolution function. The reason is to investigate the effect of the pooling layer and strided convolution on energy efficiency and test accuracy. Both neural network architectures used for MNIST are summarized in Table~\ref{mnistarchitectures}. \par

\begin{table}
	\centering
	\begin{tabular}{ | >{\centering}p{4.8cm} | >{\centering}p{5.5cm} | }
		\hline
		ConvPool-CNN & All-CNN \tabularnewline
		\hline
		\multicolumn{2}{|c|}{Input 28$\times$28 gray-scale image} \tabularnewline
		\hline
		6$\times$6 conv. 6 ReLU &   6$\times$6 conv. 6 ReLU \tabularnewline
		\hline
		5$\times$5 conv. 12 ReLU         &   \multirow{2}{*}{\vtop{\hbox{\strut 5$\times$5 conv. 12 ReLU}\hbox{\strut with stride 2}}} \tabularnewline
		\cline{1-1}
		2$\times$2 max-pooling, stride 2 &    \tabularnewline
		\hline
		4$\times$4 conv. 24 ReLU         &   \multirow{2}{*}{\vtop{\hbox{\strut 4$\times$4 conv. 24 ReLU}\hbox{\strut with stride 2}}} \tabularnewline
		\cline{1-1}
		2$\times$2 max-pooling, stride 2 &    \tabularnewline
		\hline
		\multicolumn{2}{|c|}{Fully connected layer with 200 neurons, dropout} \tabularnewline
		\hline
		\multicolumn{2}{|c|}{10-way softmax layer} \tabularnewline
		\hline
	\end{tabular}
	\caption{Model description of the two architectures for MNIST dataset.}\label{mnistarchitectures}
\end{table}

First architecture is built as a [Conv-ReLU-Conv-ReLU-Pool-Conv-ReLU-Pool-FC-Softmax] structure while second architecture is built as [Conv-ReLU-StridedConv-ReLU-StridedConv-ReLU-FC-Softmax]. The sizes of three convolutional layers and 1 fully connected layer are determined as 6, 12, 24 and 200, respectively. These configurations are determined after a large scale experimentation over the datasets. Both pooling and strided convolutional operations are used to shrink the input size by a factor of two in order to reduce the computational and statistical burden on the next layer. \par 

Filter sizes are determined heuristically. Since $5\times5$ filters are used in LeNet-5, filter sizes are selected to be close to this size. In order to preserve the input size for conventional convolutional layers, stride is chosen as 1 and zero padding is used accordingly. For strided convolutional layers, stride is 2 to decrease the height and width of the image by a factor of 2. For non-overlapping max-pooling operation, $2\times2$ filters with stride 2 is chosen. \par

The architectures which are applied to CIFAR-10 dataset are described in Table~\ref{cifar10architectures}. Since CIFAR-10 dataset has colored and high resolutions images compared to the images in MNIST, models adapted for higher capacity are preferred. Model capacity is expanded by increasing both the number of layers and the number of neurons at the hidden layers. The architecture with pooling layers is built as [Conv-ReLU-Conv-ReLU-Pool-Conv-ReLU-Conv-ReLU-Pool-FC-Softmax], while all-CNN architecture is build as [Conv-ReLU-StridedConv-ReLU-Conv-ReLU-StridedConv-ReLU-FC-Softmax]. Since we want to preserve the energy efficieny as far as possible, we use more convolutional layers, which can be modified as energy efficient layers, and only one fully-connected layer. The sizes of these 4 convolutional layers and 1 fully-connected layer are determined as 32, 32, 64, 64, and 512, respectively. The number of neurons in a layer and the filter sizes are selected empirically. A critical point in CIFAR-10 architectures is that more dropout is used due to the increased capacity. \par

\begin{table}
	\centering
	\begin{tabular}{ | >{\centering}p{5cm} | >{\centering}p{5cm} | }
		\hline
		ConvPool-CNN & All-CNN \tabularnewline
		\hline
		\multicolumn{2}{|c|}{Input 32$\times$32 RGB image} \tabularnewline
		\hline
		3$\times$3 conv. 32 ReLU &   3$\times$3 conv. 32 ReLU \tabularnewline
		\hline
		3$\times$3 conv. 32 ReLU         &   \multirow{2}{*}{\vtop{\hbox{\strut 3$\times$3 conv. 32 ReLU}\hbox{\strut with stride 2}}} \tabularnewline
		\cline{1-1}
		2$\times$2 max-pooling, stride 2 &    \tabularnewline
		\hline
		\multicolumn{2}{|c|}{Dropout} \tabularnewline
		\hline
		3$\times$3 conv. 64 ReLU &   3$\times$3 conv. 64 ReLU \tabularnewline
		\hline
		3$\times$3 conv. 64 ReLU         &   \multirow{2}{*}{\vtop{\hbox{\strut 3$\times$3 conv. 64 ReLU}\hbox{\strut with stride 2}}} \tabularnewline
		\cline{1-1}
		2$\times$2 max-pooling, stride 2 &    \tabularnewline
		\hline
		\multicolumn{2}{|c|}{Dropout} \tabularnewline
		\hline
		\multicolumn{2}{|c|}{Fully connected layer with 512 neurons, dropout} \tabularnewline
		\hline
		\multicolumn{2}{|c|}{10-way softmax layer} \tabularnewline
		\hline
	\end{tabular}
	\caption{Model description of the two architectures for CIFAR-10 dataset.}\label{cifar10architectures}
\end{table}

Note that the size of an image in the MNIST dataset is altered from $28\times28\times1$ to $32\times32\times1$ after Hadamard transform. As a result, the outputs of the BWN and HIN will not compatible in the combined model. In order to overcome this problem in the MNIST architectures, the filter size in the first convolutional layer whose input is Hadamard-transformed image is modified as $5\times5$ and zero-padding is not used. On the other hand, we will not have that issue for CIFAR-10 database. Since the width \& height of an image in CIFAR-10 is 32, a power of 2, the size will remain unchanged ($32\times32\times3$) after Hadamard transform.

\subsubsection{Network Hyperparamaters}

In order to make a fair comparison, similar methods are used for both networks and architectures. As optimization method, ADAM is selected as suggested in~\cite{rastegari2016xnor-binaryweightnetwork}. ADAM is an practical optimizer which is fairly robust to the choice of hyperparameters~\cite{goodfellow2016deep}. As regularization technique, dropout is chosen. In case of dropout, a neuron is kept active with a fixed probability of $p$ independent of other units. $p$ can be considered as a hyperparameter and it can be set to a number or determined by cross-validation. Although $p=0.5$ is a reasonable choice for hidden units, the optimal probability is usually closer to 1 than to 0.5~\cite{srivastava2014dropout}. In addition to ReLU layers in convolutional neural networks, ReLU is also used as the activation function for the fully-connected layer. ReLU is also an energy efficient function since the definition function of ReLU $f(x)=max(0,x)$ simply requires a comparison. 

Other hyperparameter settings are summarized in Table~\ref{table-hyperparameters}.

\begin{table}[H]
	\centering
	\begin{tabular}{  >{\centering}p{5cm}  >{\centering}p{2.25cm}  >{\centering}p{2.5cm}  }
		\hline
		 & MNIST & CIFAR-10 \tabularnewline
		\hline
		Weight Initialization&$W \sim \mathcal{N} \mathopen(0,0.1\mathclose)$ &Xavier init. \tabularnewline
		Bias Initialization&0.1 &0.1 \tabularnewline
		Mini-batch Size&100 &100 \tabularnewline
		Initial Learning Rate&0.003 &0.0005 \tabularnewline
		Exponential Decay Rate of Learning Rate&0.0001 & $10^{-6}$ \tabularnewline
		Probability of Retention for Dropout, $p$&0.75 & 0.75, 0.75, 0.5 \tabularnewline
		\hline
	\end{tabular}
	\caption{Hyperparameter settings for different datasets.}\label{table-hyperparameters}
\end{table}

\subsection{Experimental Results}

\begin{figure}[H]
	\centering
	\subfigure[MINIST]{
		\scalebox{0.3}{\includegraphics{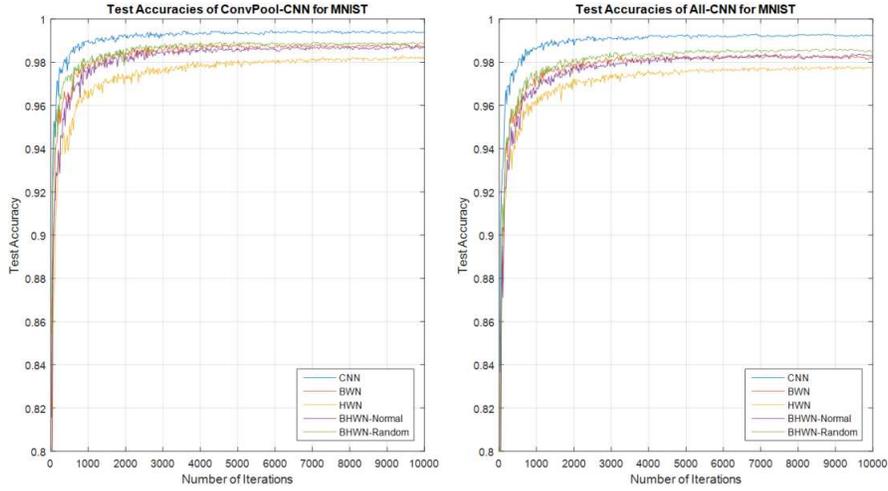}}
	}
	\subfigure[CIFAR-10]{
		\scalebox{0.3}{\includegraphics{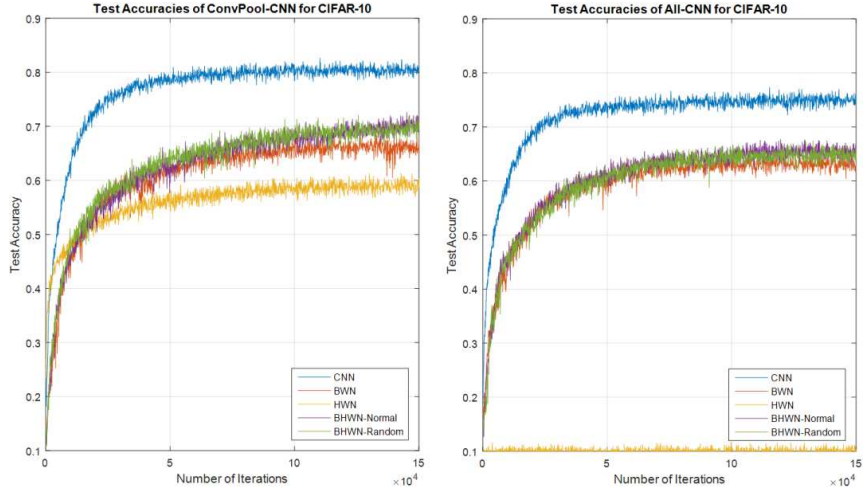}}
	}
	\caption{Test accuracy results.}
	\label{fig:test-accuracy}
\end{figure}

In the experiments, five CNN models (including the standard CNN) with two different architectures are trained on MNIST and CIFAR-10 database. These CNN models include the standard CNN, previously studied BWN and our energy efficient neural networks. The performance of the neural networks is evaluated based on test accuracies. Networks trained on MNIST dataset has 10000 iterations, while networks of CIFAR-10 dataset are trained in 150000 iterations. This means the number of epochs for MNIST and CIFAR-10 training is chosen as 17 and 300, respectively. Before learning procedure is performed, the input images are normalized to the $[0,1]$. The overall test accuracies and test accuracies corresponding to each number of iterations are presented in Table~\ref{table-testaccuracyresults} and Figure~\ref{fig:test-accuracy}, respectively.

\begin{table}[h]
	\centering
	\begin{tabular}{|p{3.5cm}|p{2cm}|p{1cm}|p{2cm}|p{1cm}|}\cline{2-5}
		\multicolumn{1}{c|}{}&\multicolumn{2}{c|}{MNIST}&\multicolumn{2}{c|}{CIFAR-10}\\\cline{2-5}
		\multicolumn{1}{c|}{}&~ConvoPool-\mbox{~~~~~~}CNN&~All-CNN&~ConvoPool-\mbox{~~~~~~}CNN&All-CNN\\\hline
		CNN&99.48&99.31&82.64&77.32\\\hline
		BWN&98.88&98.37&68.72&65.36\\\hline
		HIN&98.32&97.84&61.36&11.76\\\hline
		BWHIN-NormalAvg&98.79&98.40&72.10&67.70\\\hline
		BWHIN-RandomAvg&98.96&98.61&72.65&67.30\\\hline
	\end{tabular}
	\caption{Test accuracy results (in percentage) for CIFAR-10 dataset.}\label{table-testaccuracyresults}
\end{table}

By looking at the results presented in Table~\ref{table-testaccuracyresults} and Figure~\ref{fig:test-accuracy}, one can observe that our energy-efficient solution sacrifice the classification performance slightly in order to achieve a greater energy efficiency. While the proposed networks almost achieve the state-of-the-art result for MNIST dataset, there is an considerable accuracy gap between the state-of-the-art results and the results of the proposed networks. This problem concerning CIFAR-10 dataset can be solved by modifying the size  of the network, since binarized neural networks may need larger network structures than the networks with real-valued parameters~\cite{kim2016bitwise}. As observed from Table~\ref{table-testaccuracyresults}, it's clear that our CNN-based energy efficient neural networks with ConvPool architecture have better classification results than the All-CNN networks. Although networks with pooling layer have greater test accuracies, All-CNN can be preferred over ConvPool-CNN, since they are more energy-efficient. Here, both a convolutional layer with stride 1 and pooling layer is replaced by a convolutional layer which has stride greater than 1. Number of multiplications in the convolutional layer is decreased by a significant amount with strided convolution. While the downsampling is performed along with full convolution, a large number of values are already computed in convolutional layer, then many of these values are discarded with pooling operation. This is computationally wasteful; it will take more time and use more memory than strided convolutional layer. If we take the risk of less test accuracy in order to achieve energy efficiency, strided convolution should be used instead of pooling. There is also a significant point about All-CNN. If we increase the capacity of All-CNN by increasing the size of the layers and/or making a deeper network, image classification can be performed without the loss of accuracy. In some cases, it may even give better results~\cite{springenberg2014striving-allconvnet}. \par

Binary Weight Network is a very robust network to the changes of the hyperparameters. As seen from Table~\ref{table-testaccuracyresults}, this model could train the network for any cases. On the other hand, Hadamard-transformed Image Network is slightly worse than BWN. The reason could be the slight change in the original architecture. If Hadamard transform is applied to an image whose width \& height is not a power of 2, size of the image changes such that the width and height is increased to the next power of 2 of their original values. Hence, if we want to feed the Hadamard-transformed images to the network as input images, we may have to modify the architecture of the network. That might be the reason why HIN has worse performance and is a lossy network than BWN. Combined models work as expected; they have better test accuracies than their sub-networks BWN and HIN. When the conventional averaging has the lower test accuracies than BWN, the random averaging is always better than both BWN and BWHIN-Normal models. Hence both averaging techniques should be tried and it should be observed which one has better test accuracies even though the random-average works for most of the cases. Combined models are also a good solution to the failure of one sub-network. For example, when HIN is not trained, BWN compensates this failure and BWN-Normal/BWN-Random gives better results than both BWN and HIN. Hence it can be said that our combined model is also robust in case one of the subnetworks cannot be trained.

\section{Conclusion}

In this work, we study CNN-based enery efficient neural networks for the image classification tasks of MNIST and CIFAR-10 databases. These models are BWN, HIN, BWHIN-Normal and BWHIN-Random. As observed from these networks, energy efficiency comes with a small loss of accuracy. BWN gives satisfying results, while HIN sometimes may struggle to train the network. The combined models BWHIN-Normal and BWHIN-Random, which are our principle contributions, certainly improves the performance of their sub-networks BWN and HIN. Hence as an energy efficient model, the combined model can be preferred with its superior classification performance. \par

As future work, proposed energy efficient neural networks can be used for the classification of other datasets such as CIFAR-100, Street View House Numbers (SVHN) dataset or ImageNet. The networks can be also implemented with bigger capacity by increasing the size of a hidden layer or with more hidden layers, i.e. as a deeper network. Effects of data preprocessing or usage of unsupervised learning as pretraining on energy efficient NN might be another research topic.

\bibliographystyle{ieeetr}

\end{document}